\pdfoutput=1

\documentclass[11pt]{article}

\usepackage{acl2023}

\usepackage{times}
\usepackage{latexsym}
\usepackage[T1]{fontenc}
\usepackage[utf8]{inputenc}
\usepackage{microtype}
\usepackage{inconsolata}
\usepackage{enumitem}

\usepackage{multirow}
\usepackage{hyperref}
\usepackage{graphicx}
\usepackage{subcaption}
\usepackage{adjustbox}
\usepackage{caption}

\AtBeginDocument{%
  \providecommand\BibTeX{{%
    \normalfont B\kern-0.5em{\scshape i\kern-0.25em b}\kern-0.8em\TeX}}}

\title{The Daunting Dilemma with Sentence Encoders: Success on Standard Benchmarks, Failure in Capturing Basic Semantic Properties}

\author{Yash Mahajan , Naman Bansal,
Shubhra Kanti Karmaker ("Santu") \\
Big Data Intelligence (BDI) Lab, Auburn University, Alabama, USA\\
\{yzm0034, nbansal sks0086\}@auburn.edu}

\begin{document}

\maketitle

\begin{abstract}

    In this paper, we adopted a retrospective approach to examine and compare five existing popular sentence encoders, i.e., Sentence-BERT, Universal Sentence Encoder (USE), LASER, InferSent, and Doc2vec, in terms of their performance on downstream tasks versus their capability to capture basic semantic properties. Initially, we evaluated all five sentence encoders on the popular SentEval benchmark and found that multiple sentence encoders perform quite well on a variety of popular downstream tasks. However, being unable to find a single winner in all cases, we designed further experiments to gain a deeper understanding of their behavior. Specifically, we proposed four semantic evaluation criteria, i.e., Paraphrasing, Synonym Replacement, Antonym Replacement, and Sentence Jumbling, and evaluated the same five sentence encoders using these criteria. We found that the Sentence-Bert and USE models pass the paraphrasing criterion, with SBERT being the superior between the two. LASER dominates in the case of the synonym replacement criterion. Interestingly, all the sentence encoders failed the antonym replacement and jumbling criteria. These results suggest that although these popular sentence encoders perform quite well on the SentEval benchmark, they still struggle to capture some basic semantic properties, thus, posing a daunting dilemma in NLP research.
     
\end{abstract}

\section{Introduction}\label{intro}

One of the fundamental tasks in NLP is to map sentences computationally into dense vector representations for subsequent analysis. These dense vectors of fixed size, which are known as ``sentence embeddings'', represent the meaning of sentences in some latent semantic space. Till today, many supervised \cite{conneau-etal-2017-supervised} and unsupervised \cite{le2014distributed} methods have been proposed to learn embeddings for a given sentence. For instance, Doc2vec \cite{le2014distributed}, proposed in 2014, is one of the earliest sentence encoding techniques that uses a deep neural network. In 2017, InferSent \cite{conneau-etal-2017-supervised}, developed by Facebook, used Bi-LSTM networks to learn sentence embeddings. Later, in 2017, Transformers~\cite{vaswani2017attention} were introduced and subsequently, many transformer-based sentence encoders have been proposed since then including  BERT \cite{devlin-etal-2019-bert}, USE (Universal-Sentence-Encoder) \cite{cer2018universal}, Sentence-BERT~\cite{reimers-gurevych-2019-sentence}, LASER~\cite{artetxe-schwenk-2019-margin} etc..

\begin{table*}[!htb]
\centering
    \begin{adjustbox}{width=0.85\linewidth}
    
    \begin{tabular}{r|l|l}
        \multicolumn{3}{c} {Original Sentence: ``\textit{Levin's attorney, Bo Hitchcock, declined to comment last Friday}''}\\\hline
        \textbf{Perturbation Task} & \textbf{Example Sentence} & \textbf{Expected Encoding}\\\hline
        
        \textbf{Paraphrasing} &{Hitchcock has declined to comment on the case, as has Levin.} & Similar to Original\\\hline
        \textbf{Synonym Replacement} & {Levin's attorney, Bo Hitchcock, \textit{refused} to comment last Friday.} & Similar to Original\\\hline
        \textbf{Antonym Replacement}&{Levin's attorney, Bo Hitchcock, \textit{accepted} to comment last Friday.} & Diverse from Original \\\hline
        \textbf{Sentence Jumbling }&{Levin's attorney \textit{to} Bo Hitchcock, declined\textit{,} comment last Friday.} & Diverse from Original\\\hline
    \end{tabular}
    \end{adjustbox}
    \vspace{-2mm}
    \caption{Example of four unsupervised Semantic Understanding Task.}
    \label{tab:samples}
\end{table*}

   

While some of these powerful sentence encoders have demonstrated superior performance on standard benchmarks and downstream NLP tasks~\cite{choi2021evaluation,conneau2018senteval}, we still lack a good understanding of the pros and cons of using different sentence encoders for any task~\cite{pham2021out}. In other words, despite achieving high accuracy numbers on benchmark datasets, it is still unclear whether they indeed capture basic linguistic properties (which is desired) while doing so. To investigate this in detail, we adopt a retrospective approach in this paper to analyze and compare five existing popular sentence encoders in terms of their capability to capture the basic semantics. Specifically, we designed four semantic criteria to evaluate sentence encoders:  1) Paraphrasing, 2) Synonym Replacement, 3) Antonym Replacement, and 4) Sentence Jumbling, as shown in Table~\ref{tab:samples}, to quantify how well a sentence encoder can capture the semantic relations between two related sentences.

Computationally, given a sentence $S$ and its corresponding embedding $S_x$, the basic idea here is to perturb $S$ according to a particular criterion to create $S'$ (with embedding $S'_x$), then look at how similar/different two embedding vectors $S_x$ and $S'_x$ are and match those observations against the expected behavior. For example (see Table~\ref{tab:samples}), given original sentence: ``\textit{Levin's attorney, Bo Hitchcock, declined to comment last Friday}'', an example of synonym replacement perturbation is: ``Levin's attorney, Bo Hitchcock, \textit{refused} to comment last Friday''. Obviously, these two sentences are very similar, and intuitively, a good sentence encoder should produce very similar sentence embeddings for them. On the contrary, an Antonym Replacement or Sentence Jumbling perturbation usually shifts the meaning of the sentence significantly, and therefore, a good sentence encoder should yield a somewhat diverse embedding for Antonym Replacement/Jumbling.

Based on the intuitions mentioned above, we designed experiments to test five popular sentence encoders with respect to our four semantic evaluation criteria. Note that these four criteria only constitute a subset of linguistic properties that we argue a good sentence encoder should hold, \textit{but it is far from an exhaustive list, which is beyond the scope of this paper}. A benefit of these four evaluation criteria is that they can be experimentally evaluated in an unsupervised fashion without requiring a specific downstream task.

For experiments, we initially evaluated and analyzed the performance of five popular sentence encoders, i.e., Sentence-BERT, Universal Sentence Encoder (USE), LASER, InferSent, and Doc2vec, on the SentEval benchmark and found that there is no single winner for all benchmark tasks. Next, we conducted extensive experiments to test our four semantic evaluation criteria on these five encoder models, and the Sentence-BERT model demonstrated the closest to expected behavior in the case of the Paraphrasing criterion, while LASER performed the best in the case of the Synonym Replacement criterion. On the contrary, all the sentence encoders failed to satisfy the Antonym Replacement and Sentence Jumbling criteria. These results suggest that a sentence encoder model can perform quite well on the SentEval benchmark even though they fail to satisfy some of the basic semantic evaluation criteria. This result raises several daunting philosophical dilemmas in NLP research in general, e.g., when can we claim a sentence encoder as ``good'' vs. ``bad''? Do we only care about the performance of downstream tasks even when basic linguistic properties are violated? Is the SentEval benchmark challenging enough, or do we need to add harder tasks into the SentEval benchmark that demands a deeper semantic understanding for a more accurate evaluation of sentence encoders? We urge the community to conduct further research on these questions based on our study.

\section{Related Works}\label{LR}
By far, many have proposed a variety of techniques to generate the embedding for a given sentence. Doc2Vec \cite{le2014distributed} is an unsupervised technique that generates embedding based on the variable-length piece of text and creates unique embeddings for each paragraph in a document. Later, others attempted to learn sentence embedding using auto-encoder \cite{socher2011dynamic,hill2016learning}, \cite{hu2017toward}. On the other hand, InferSent \cite{conneau-etal-2017-supervised} used SNLI~\cite{dolan2004unsupervised} and Multi-genre NLI labeled data~\cite{williams2017broad} and learned the sentence embedding using the Bi-LSTM with max-pooling architecture and a Siamese network.

More recently, \citet{cer2018universal} proposed ``Universal Sentence Encoder'' (USE), which is trained on the combination of supervised and unsupervised NLI (Natural Language Inference) data, and it has effectively produced sophisticated sentence embeddings. Sentence BERT (SBert) \cite{reimers-gurevych-2019-sentence}, which is trained on Wikipedia corpus and news-wire articles and later fine-tuned on SNLI and Multi-Genre NLI dataset. These models have been trained rigorously on a large corpus of data, and many of them used data parallelisms \cite{wieting2017paranmt,artetxe2019massively,wieting2019simple,wieting2019bilingual}, natural language inference (NLI) \cite{conneau2017supervised,conneau2018you,reimers2019sentence}, or a combination of both \cite{subramanian2018learning}.

However, recently \citet{reimers-gurevych-2019-sentence,li2020sentence, pham2021out} reported that these pre-trained language models produce poor embeddings for semantic similarity tasks. Many pre-trained language models are designed for task-specific purposes; as a result, the embeddings generated by the models could be biased. To further investigate this issue in this paper, we conduct a systematic study of popular sentence encoders by proposing four basic semantic evaluation criteria and report our findings to inform the research community.

\section{Evaluation on SentEval Benchmarks}\label{senteval}

\begin{table*}[ht]\tiny
    \centering
    \begin{adjustbox}{width=0.9\linewidth}
    \begin{tabular}{c|c|c|c|c|c|c|c|c}\hline
         \textbf{Model} & \textbf{MR}&\textbf{CR}&\textbf{SUBJ}&\textbf{MPQA}&\textbf{SSTb}&\textbf{TREC}&\textbf{MRPC}&\textbf{Avg} \\\hline \hline
         \textbf{SBERT}& \textbf{83.95}&\textbf{88.98}&\textbf{93.77}&\textbf{89.51}&\textbf{90.01}&84.8&76.28&\textbf{86.9}\\
         \textbf{USE}& 75.58&81.83&91.87&87.17&85.68&\textbf{92.2}&69.62&83.42\\
         \textbf{Infersent}&81.1 &86.3&92.4&90.2&84.6&88.2&76.2&85.57\\
         \textbf{LASER}&56.14 &63.89&67.65&72.36&72.85&79.85&\textbf{89.19}&72.04\\
         \textbf{Doc2Vec}& 49.76&63.76&49.16&68.77&49.92&19.2&66.49&52.43\\\hline

    \end{tabular}
    \end{adjustbox}   
    \caption{Evaluation of existing sentence encoders on SentEval Benchmark. The accuracy scores are generated using the SentEval toolkit on different classification tasks. The scores are generated using 10-fold cross-validation.}
    \label{tab:senteval-tab}
\end{table*}

SentEval~\cite{conneau2018senteval} is a widely used framework for evaluating the efficacy of sentence embeddings. Here, sentence embeddings are used to perform various classification tasks. Specifically, the SentEval toolkit uses a logistic regression classifier or multi-layered perceptron (MLP), which deploys a 10-fold cross-validation methodology across a range of classification tasks. The testing fold is then utilized to compute the prediction accuracy of the classifiers.

In this work, we assess the effectiveness of five distinct sentence encoders on seven datasets from the SentEval benchmark to identify the best one. 
\begin{enumerate}[leftmargin=*,itemsep=0ex,partopsep=0.5ex,parsep=0ex]
    \item \textbf{MR}: Movie review dataset for sentiment binary classification task~\cite{MR}. 
    \item \textbf{CR}: Sentiment prediction on Product review dataset with binary labels~\cite{CR}.
    \item \textbf{MPQA}: An opinion polarity dataset with binary classification task~\cite{MPQA}.
    \item \textbf{SSTb}: Stanford Sentiment Treebank dataset with binary labels~\cite{SST}.
    \item \textbf{SUBJ}: Subjective prediction from movie reviews/plot summaries~\cite{SUBJ}. 
    \item \textbf{TREC}: Fine-grained question-type classification grom TREC~\cite{TREC}.
    \item \textbf{MRPC}: Mircosoft Paraphrase Corpus from parallel news sources~\cite{MRPC}. 
\end{enumerate}

The accuracy scores of each sentence encoder can be found in Table ~\ref{tab:senteval-tab}. The SBERT model exhibits superior performance, generating more useful embeddings than other models on five out of seven benchmark datasets, with the highest average performance of 86.9. However, USE and Infersent model also demonstrates very similar performance to SBERT with nearly one and three-point difference, respectively, in terms of average scores. This raises concerns regarding the best sentence encoder to use for an unseen task, as we do not want an encoder that generates quality embeddings for a particular task but fails to capture the context in other tasks. For instance, the SBERT model performed poorly on the MRPC dataset and TREC dataset but performed well on other benchmarks. Thus, raising doubts about the encoder's ability to provide quality embeddings for any task.  Therefore, we need to investigate in-depth whether a sentence encoder can differentiate between two orthogonal and non-orthogonal sentences. Thus, we designed and created four intuitive semantic criteria \textit{Paraphrase, Synonym Replacement, Paraphrase Vs. Antonym Replacement and, Paraphrase Vs. Jumbled Sentence} that are simple yet important to evaluate sentence encoder (see section~\ref{sec:hypo}). The evaluation of these criteria will provide detailed insight into how sentence encoders understand the natural language and how efficiently they capture the context in their embeddings. 

\section{Four Semantic Evaluation Criteria}\label{sec:hypo}

\begin{enumerate}[leftmargin=*,itemsep=1ex,partopsep=0ex,parsep=0ex]
    \item \textbf{Criterion-1 (Paraphrasing)}: As our first criterion, we argue that: ``A \textit{good} sentence encoder should generate similar embeddings for two sentences which are paraphrases of each other''. Similarly, a \textit{good} sentence encoder should generate significantly different embeddings for two unrelated sentences. Therefore, the difference between the average similarity score (in terms of sentence embeddings) of a collection of paraphrase pairs and that of non-paraphrase pairs should be high for a \textit{``good''} sentence encoder. 

    \item \textbf{Criterion-2 (Synonym Replacement)}: For the second criterion, we argue that: ``If we replace $n$ words (where, $n$ is small) from sentence $S$ with their respective synonyms to create another sentence $S'_P$, a good sentence encoder will yield similar embeddings for $S$ and $S'_P$ in the latent semantic space''. The intuition here is that synonym replacement does not alter the meaning of a sentence significantly, hence, the embeddings are also expected to remain similar. 

    \item \textbf{Criterion-3 (Paraphrase Vs. Antonym Replacement)}: For the third criterion, we argue that: ``Given a sentence $S$, its paraphrase $S'_P$ and an antonym-replaced sentence $S'_A$ (created by replacing exactly one word (verb/adjective) in $S$ with its antonym), $S'_P$ should be semantically more similar to $S$ than $S_A'$ to $S$ by some clear margin, i.e., $Sim(S,S'_P) - Sim(S,S_A') > \epsilon_1$, where $\epsilon_1$ denotes the expected minimum margin. The intuition here is that a good sentence encoder should generate embeddings in a manner such that any paraphrase is closer to the original sentence than an antonym-replaced sentence in the latent semantic space. This will ensure that the encoding can indeed differentiate between paraphrased and antonym-replaced sentences and capture their semantic variance.

    \item \textbf{Criterion-4 (Paraphrase Vs. Sentence Jumbling)}: Our fourth criterion is similar to the third one, except that instead of antonym replacement, we consider sentence jumbling. Formally, Given a sentence $S$, its paraphrase $S'_P$ and a jumbled sentence $S'_J$ (created by randomly swapping $n$ pairs of words among each other in the original sentence $S$), $S'_P$ should be semantically more similar to $S$ compared to the same for $S_J'$ by some clear margin, i.e, $Sim(S,S'_P) - Sim(S,S_J') > \epsilon_2$, where $\epsilon_2$ denotes the expected minimum margin. The intuition here is that a good sentence encoder should generate embeddings in a manner such that any paraphrase of a given sentence should be closer to the original sentence than a jumbled one in the latent semantic space.

\end{enumerate}

\section{Experiments}\label{exp_setup}

\subsection{Data-set}\label{dataset}
In this work, we used three publicly available paraphrasing data-sets with human-annotated labels. All three datasets come with binary labels assigned to each pair of sentences. Label $1$ (Pos) indicates that the pair of sentences have a similar meaning and $0$ (Neg) indicates otherwise. The data sets are \textbf{1) QQP} (Quora Questions Pair) dataset \cite{chen2018quora}, which is the collection of paraphrased and non-paraphrased pairs of questions. \textbf{2) PAWS-WIKI} (Paraphrase Adversaries from Word Scrambling-Wikipedia) dataset \cite{zhang2019paws}, which is a collection of pair of sentences from Wikipedia with high lexical overlaps. These pairs of sentences are labeled with 1's and 0's for paraphrase and non-paraphrase pairs, respectively. And, \textbf{3) MRPC} (Microsoft Research Paraphrasing Corpus) dataset \cite{dolan-brockett-2005-automatically}, which is a collection of sentence pairs extracted from news articles. More details can be found in appendix~\ref{appendix_dataset}.

\subsection{Sentence Encoder Models}\label{sec:models}
In this work, we compared five popular sentence embedding methods:
1) Universal Sentence Encoder (USE) \cite{cer2018universal}, 2) Sentence-BERT (SBert) \cite{reimers-gurevych-2019-sentence}, 3) InferSent \cite{conneau-etal-2017-supervised}, 4) Language-Agnostic-SEntence Representation (LASER)~\cite{artetxe-schwenk-2019-margin}, and 5) Document To Vector (Doc2Vec or D2V) \cite{le2014distributed}. Detailed descriptions of the models can be found in appendix~\ref{secc:m_para}.

\section{Results}\label{results}

All models were evaluated on the four proposed criteria. Here, we argue that \textit{a sentence encoder shall pass all the criteria in order to be considered a ``good'' sentence encoder.} Our results on the three datasets (refer to section~\ref{dataset}) are described below. All the evaluations were performed on a Google Colab GPU server and a local system having an Intel i5 processor and 8GB of RAM.

\begin{enumerate}[leftmargin=*,itemsep=1ex,partopsep=0.5ex,parsep=0ex]
    \item \textbf{Criterion-1 (Paraphrasing)}: To evaluate criterion-1, we took sentence pairs (both paraphrases and non-paraphrases) from each data set with no modification and encoded each sentence using the five popular sentence encoders mentioned in Section~\ref{sec:models}. Next, we computed  \textit{Cosine Similarity} on the embedding vector pairs and calculated the similarity between the sentence pairs, and further averaged the similarity scores in case of paraphrases (positive instances) and non-paraphrases (negative instances), separately. Finally, we computed the difference between the average similarity of paraphrase and non-paraphrase pairs to evaluate our criterion. As mentioned in section~\ref{sec:hypo}, this difference is expected to be high in the case of a ``good'' sentence encoder. 

\begin{table}[!htb]\small
    \centering
    \begin{adjustbox}{width=\linewidth}
    \begin{tabular}{lp{.45cm}p{.45cm}p{.65cm}p{0.76cm}p{.95cm}p{.85cm}}
    \hline
    \textbf{Dataset} &\textbf{} & \textbf{USE} & \textbf{SBert} & \textbf{Infer-Sent} & \textbf{LASER} & \textbf{D2V} \\\hline\hline
    \multirow{3}{*}{QQP}& Pos.  & 0.83 & 0.87 & 0.86 & 0.83 & 0.34 \\
    &Neg.  & 0.58 & 0.56  & 0.77 & 0.71 & 0.32 \\
    &Diff.  & 0.25 & \textbf{0.31} & 0.09 & 0.12 & 0.02\\\hline
    
    \multirow{3}{*}{PAWS}&Pos.  & 0.95 & 0.97 & 0.96 & 0.97 & 0.70 \\
    &Neg.  & 0.94 & 0.96  & 0.96 & 0.94 & 0.73 \\
    &Diff.  & 0.01 & 0.01 & 0.007 & 0.03 & -0.03\\\hline
    
    \multirow{3}{*}{MRPC} &Pos.  & 0.78 & 0.83 & 0.91 & 0.87 & 0.60 \\
    &Neg.  & 0.67 & 0.68 & 0.88 & 0.81 & 0.50 \\
    &Diff.  & 0.11	&\textbf{0.15 }	&0.03	&0.06	&0.10\\\hline
    \end{tabular}
    \end{adjustbox}
    \caption{Average Cosine Similarity for Hypothesis-1 Paraphrasing task. Here, Positive (\textbf{Pos.}) and Negative (\textbf{Neg.}) means paraphrase-pairs and non-paraphrase-pairs, respectively.} 
    \label{tab:para-exp-1}
\end{table}

    \smallskip
    Table \ref{tab:para-exp-1} summarizes the results for criterion-1. Overall, the SBert model was able to better distinguish between 
    paraphrase (positive) and non-paraphrase (negative) pairs for QQP and MRPC data sets, followed by Universal Sentence Encoder (USE) as the second best. Whereas, all encoders failed to differentiate in the case of the PAWS-WIKI data set (Note that, sentence pairs in the PAWS-WIKI data-set share high lexical overlap, hence, it is easy for encoders to get confused). In fact, Doc2Vec (D2V) failed to differentiate in the case of all three data sets, while InferSent and LASER showed sub-optimal performance. The performance of all models aligned with the SentEval benchmark (refer ~\ref{tab:senteval-tab}). Hence, we conclude that the SBERT and USE models pass this criterion, and the rest of the  models struggle. Also, the SBERT model was the best-performing among all models.

    \medskip
    \textbf{Criterion-1 (Alternative Setup)}: In this criterion, we aimed to test an alternative setup to evaluate criterion-1. Instead of selecting negative pairs that are non-paraphrases yet somewhat related, we created negative pairs by randomly sampling two sentences, each belonging to a different topic. The idea behind this is to create negative pairs from orthogonal topics with fewer overlapping words, leading to a lower expected similarity between them. In contrast, positive samples remained the same as in the original setup of criterion-1 evaluation.
    
\begin{table}[!htb]\small
    \centering
    
    \begin{adjustbox}{width=\linewidth}
    \begin{tabular}{lp{.45cm}p{.45cm}p{.65cm}p{0.76cm}p{.95cm}p{.85cm}}
    \hline
    \textbf{Data-set} &\textbf{} & \textbf{USE} & \textbf{SBert} & \textbf{Infer-Sent} & \textbf{LASER} & \textbf{D2V} \\\hline\hline
    \multirow{3}{0.08\linewidth}{QQP}&Pos.& 0.83 & 0.87 & 0.86 & 0.83 & 0.34 \\
    &Neg. & 0.13 & 0.03 & 0.58 & 0.50 & 0.24 \\
    &Diff.  & 0.70 & \textbf{0.84} & 0.28 & 0.33 & 0.10\\\hline
    
    \multirow{3}{*}{PAWS}&Pos. & 0.95 & 0.97 & 0.96 & 0.97 & 0.70 \\
    &Neg.  & 0.08 & 0.01  & 0.63 & 0.61 & 0.18 \\
    &Diff.  & 0.87 & \textbf{0.96} & 0.33 & 0.36 & 0.52\\\hline
    
    \multirow{3}{*}{MRPC}&Pos.  & 0.78 & 0.83 & 0.91 & 0.87 & 0.60 \\
    &Neg. & 0.07 & 0.03 & 0.68 & 0.57 & 0.36 \\
    &Diff.  & 0.71	&\textbf{0.80}	&0.23	&0.30	&0.24\\\hline
    
    \end{tabular}
    \end{adjustbox}
    \caption{Average Cosine Similarity for Hypothesis-1 Paraphrasing task (Alternative Setup).}
    \label{tab:para-exp-2}
\end{table}

    
    
    

    \smallskip
    Table \ref{tab:para-exp-2} summarizes the results for criterion-1 (Alternative Setup). Again, the SBert model was able to best distinguish between paraphrase (positive) and non-paraphrase (negative) pairs for QQP and MRPC data sets, followed by Universal Sentence Encoder (USE) as the second best. This time, the differences are much larger. Consistently with SentEval (refer to Table~\ref{tab:senteval-tab} trend and Table \ref{tab:para-exp-1}, InferSent, LASER, and Doc2Vec were again found to be sub-optimal.

    \item \textbf{Criterion-2 (Synonym Replacement)}: To create the synonyms perturbed sentence, we first randomly chose $n$ (n=1,2,3) words that are verb/adjective (not stop-words) and replace them with the synonyms that are generated by \textit{WordNET} toolkit~\cite{miller1995wordnet}. Next, after replacing $n$ words with their synonym(s), we encoded both sentences (original and perturbed) using the five-sentence encoder techniques and computed their cosine similarities. This process was repeated for all three datasets, and Table~\ref{tab:syn_similarity-1} reports the average of these cosine similarity scores. As mentioned in Section~\ref{sec:hypo}, this similarity is expected to be high in the case of a ``good'' sentence encoder to satisfy criterion-2.

\begin{table}[!htb]\small
    \centering
    \begin{adjustbox}{width=\linewidth}
    \begin{tabular}{lp{.85cm}p{.55cm}p{.65cm}p{0.76cm}p{.95cm}p{.85cm}}
    \hline
     {} &\textbf{Data-set} &\textbf{USE} & \textbf{SBert} & \textbf{Infer-Sent} & \textbf{LASER} & \textbf{D2V} \\\hline\hline
    \multirow{3}{*}{n=1}&QQP& 0.895 & 0.916 & 0.938 & \textbf{0.948} & 0.684 \\
    &WIKI. & 0.951 &0.964 & 0.969 & \textbf{0.982} & 0.769 \\
    &MPRC  & 0.949 & 0.948 & 0.975 & \textbf{0.978} & 0.795\\\hline 
    
    \multirow{3}{*}{n=2} & QQP & 0.809 & 0.848 & 0.895 & \textbf{0.907} & 0.587 \\ 
    &WIKI. & 0.902 & 0.928 &0.943 & \textbf{0.966} & 0.699\\
    &MPRC  & 0.900 & 0.897 & 0.955 & \textbf{0.961} & 0.671\\\hline
    
    \multirow{3}{*}{n=3}&QQP& 0.739 & 0.791 & 0.865 & \textbf{0.879} & 0.528 \\ 
    &WIKI. &0.857 & 0.892  & 0.919 & \textbf{0.950} & 0.670 \\
    &MPRC  & 0.850 & 0.846 & 0.935 & \textbf{0.944} & 0.584\\\hline 
    
    \end{tabular}
    \end{adjustbox}
    \caption{The Average Cosine Similarity between the Original and the Synonym Replaced Sentence. Here $n=\{1,2,3\}$ represent the number times the words are replaced in a sentence.}
    \label{tab:syn_similarity-1}
\end{table}


    
    From Table~\ref{tab:syn_similarity-1}, we notice that the LASER and Infersent models captured higher cosine similarity between the original and perturbed sentences than any other model for all n's. However, LASER and Infersent were sub-optimal in the case of the criterion-1 (Paraphrasing) task. On the other hand, similarity scores yielded by SBert and USE for Synonym Replacement are pretty close to the same for LASER and Infersent, and hence, SBert and USE still remain the better choice considering both criteria 1 and 2. Also, the similarity score drops gradually with an increase in the order of \textit{n}, i.e., an increase in word replacement for each sentence. Hence, we can say that except for the D2V model, all other models satisfy criterion-2.

\begin{figure}[!pt]
    \centering
    \begin{subfigure}[b]{0.48\textwidth}
        \includegraphics[width=\textwidth]{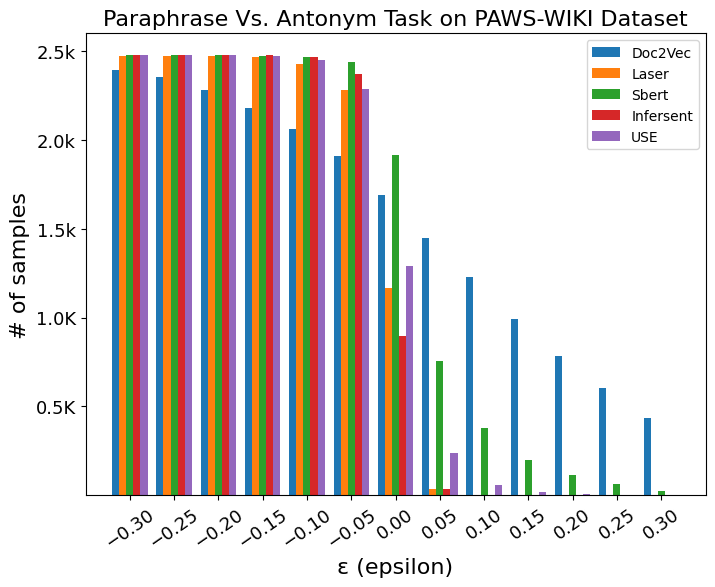}
        \caption{Criterion-3 evaluation on PAWS-WIKI dataset.}
        \vspace{2.5mm}
        \label{fig:paw_anto}
    \end{subfigure}
    \begin{subfigure}[b]{0.48\textwidth}
        \includegraphics[width=\textwidth]{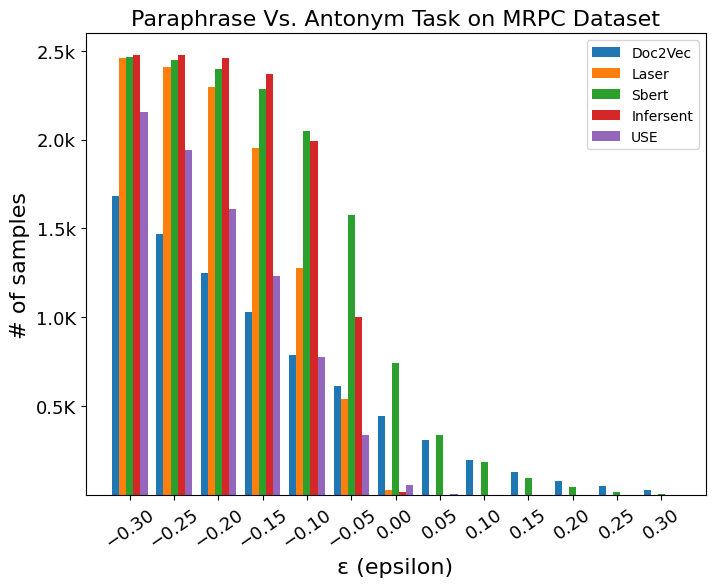}
        \caption{Criterion-3 evaluation on MRPC dataset.}
        \vspace{2.5mm}
        \label{fig:mrpc_anto}
    \end{subfigure}
    \begin{subfigure}[b]{0.48\textwidth}
        \includegraphics[width=\textwidth]{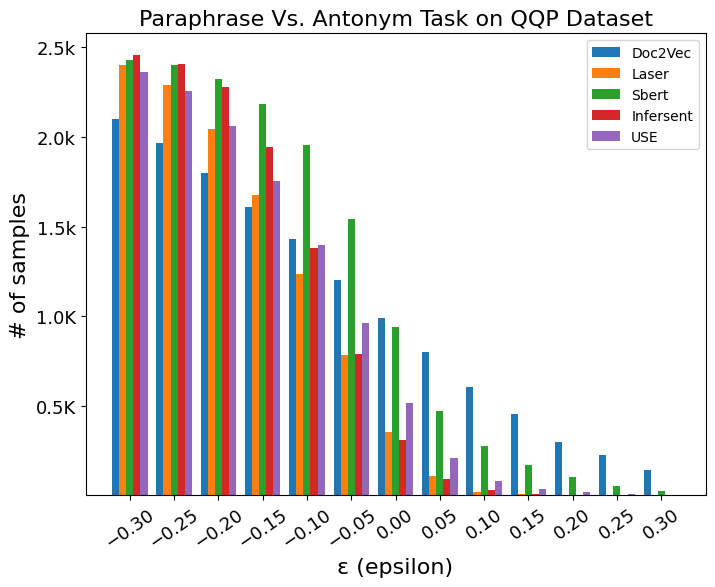}
        \caption{Criterion-3 evaluation on QQP dataset.}
        \vspace{2.5mm}
        \label{fig:qqp_anto}
    \end{subfigure}
    \caption{The figures demonstrate the cosine similarity difference for Paraphrased Vs. Antonym hypothesis. The score are calculated based on $Sim(S,S'_P) - Sim(S,S_A') > \epsilon_1$. All five different sentence encoders were tested on all three datasets. Here the epsilon is equal to $\epsilon_1$.}
    \label{fig:anto}
\end{figure}

    \item \textbf{Criterion-3 (Paraphrase Vs. Antonym 
    Replacement)}: In the third criterion, we expect that the  paraphrased sentence $S'_P$ should be semantically closer to the original sentence $S$ compared to an antonym sentence $S'_A$. To test this criterion, we computed the cosine similarities between the sentence pairs ($S$, $S'_P$(paraphrase)) and ($S$, $S'_A$ (antonym)) separately and then, computed the difference between these two similarity scores. We used WordNet toolkit~\cite{miller1995wordnet} to create an antonym sentence and repeated this process for each sentence $S$ in a data set and, finally, plotted a cumulative histogram where the bins in the x-axis represent the expected minimum difference margins ($\epsilon_1$ = [-0.3 to 0.3]), and the y-axis represents the number of sentences $S$ in the dataset with that minimum margin, i.e., $Sim(S,S'_P) - Sim(S,S_A') > \epsilon_1$, for each sentence embedding technique (Figure \ref{fig:anto}). To satisfy the criterion, we expect that a good sentence encoder should yield higher similarity for ($S$,$S'_P$) pair than the ($S$,$S'_A$) pair, and hence, their difference in similarity score should be somewhat significant ($>>0$).
    
    \smallskip
    Upon closer examination of Figure~\ref{fig:anto}, it becomes apparent that all five sentence encoders produce left-skewed cumulative histograms, indicating that they are unable to differentiate between $S'_A$ and $S'_P$ in terms of their differences from the original sentence $S$. This observation essentially means that all five sentence encoders fail to satisfy criterion 3, as most of the samples fall within the $\epsilon_1$ range of -0.3 to 0, with very few samples having positive differences. Thus, the sentence encoders tend to produce embeddings that place $S'_A$ closer to $S$ than $S'_P$ in the latent semantic space, which is the opposite of what we expected. Surprisingly, the Doc2Vec model in Figure~\ref{fig:anto} demonstrates some positive difference between the sentence pair (($S$, $S'_P$(paraphrase)) and ($S$, $S'_A$ (antonym)). Moreover, for the PAWS-WIKI dataset (figure ~\ref{fig:paw_anto}), where all other models failed miserably, the Doc2Vec model exhibited a significantly higher positive difference between pairs. The cause of this difference is unclear, and we plan to investigate it in more detail in future work. Further, comparing these results to those of SentEval (see Table~\ref{tab:senteval-tab}) reveals an interesting finding. Four out of five models achieved relatively high accuracy scores on downstream tasks, yet all models failed to capture a desired basic linguistic property. These findings also raise questions about whether SentEval is a hard-enough benchmark for testing sentence encoders or whether we may be overly reliant on metrics such as cosine similarity score, whose underlying working is unclear yet considered as a potential similarity metric to evaluate our sentence encoders.

\begin{figure}[!pt]
    \centering
    \begin{subfigure}[b]{0.48\textwidth}
        \includegraphics[width=\textwidth]{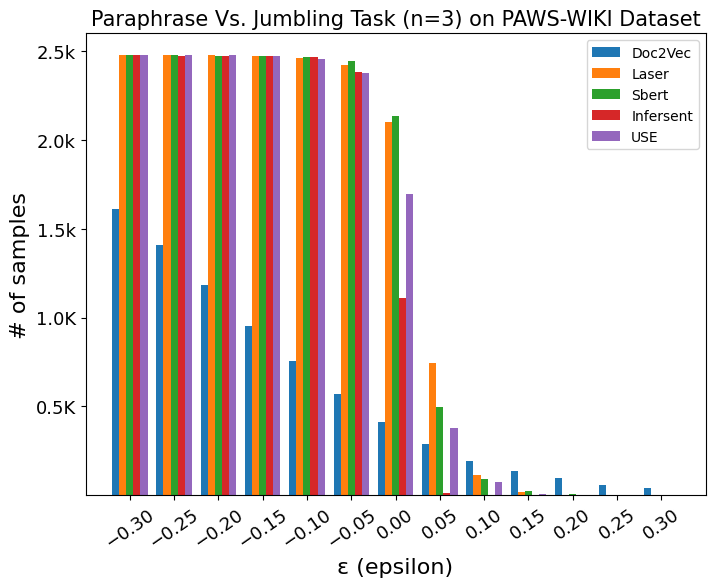}
        \caption{Criterion-4 evaluation on PAWS-WIKI dataset}
        \vspace{2.5mm}
        \label{fig:jumbling_subfig1}
    \end{subfigure}
    
    \begin{subfigure}[b]{0.47\textwidth}
        \includegraphics[width=\textwidth]{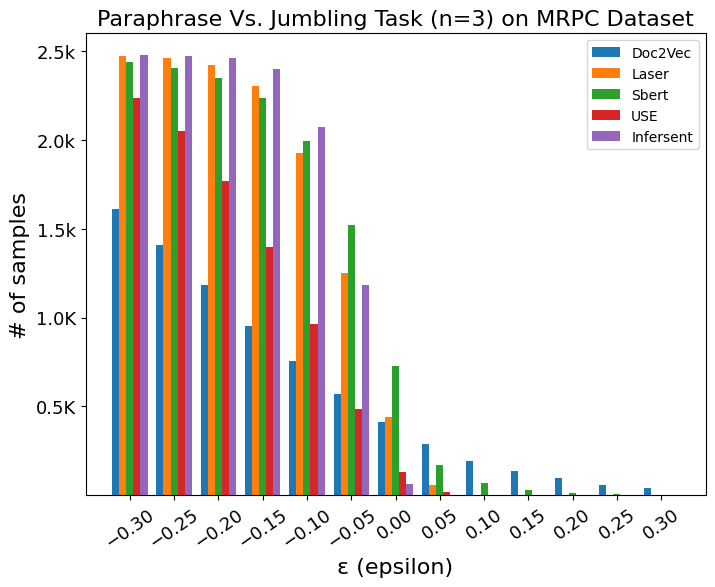}
        \caption{Criterion-4 evaluation on MRPC dataset}
        \vspace{2.5mm}
        \label{fig:jumbling_subfig2}
    \end{subfigure}
    \vspace{2.5mm}
    \begin{subfigure}[b]{0.47\textwidth}
        \includegraphics[width=\textwidth]{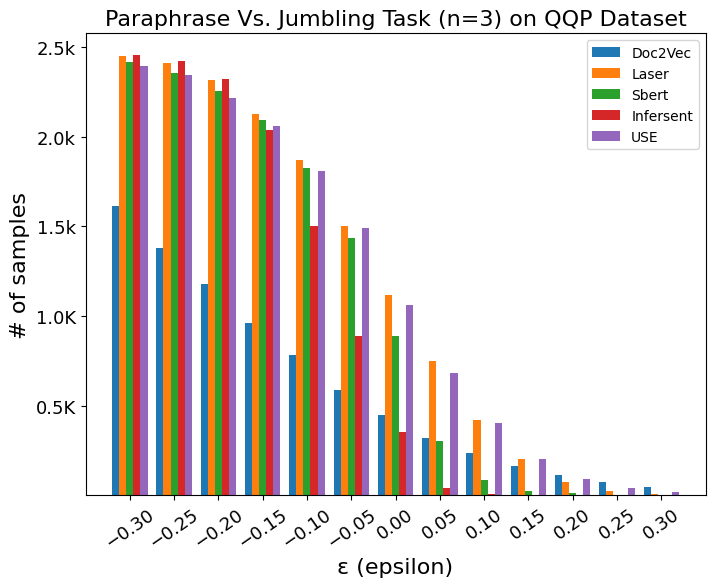}
        \caption{Criterion-4 evaluation on QQP dataset}
        \label{fig:jumbling_subfig3}
    \end{subfigure}
    \caption{The figures demonstrate the cosine similarity difference for Paraphrased Vs. Jumbling criterion. The score are calculated based on $Sim(S,S'_P) - Sim(S,S_J') > \epsilon_2$. All five different sentence encoders were tested on all three datasets. Here the epsilon is equal to $\epsilon_2$.}
    \label{fig:jumbling}
\end{figure}

    \item \textbf{Criterion-4 (Paraphrase Vs. Sentence Jumbling)}: In this criterion, we expect that when some words are swapped by each other in a sentence, the meaning of the perturbed sentence should be completely destroyed. The perturbed jumbled sentence $S'_J$ should no more convey the same meaning as the original sentence $S$, and thus, it should not be placed close to the original sentence $S$ in latent semantic space. On the other hand, a paraphrased sentence $S'_P$ conveys the same meaning and hence, should be closer to the original sentence $S$ in the same latent space. To investigate this criterion, a similar equation is used as in criterion 3, $Sim(S,S'_P) - Sim(S,S_J') > \epsilon_2$) where the similarity score of antonym $Sim(S,S'_A)$ is replaced with the similarity of a jumbled sentence, i.e., $Sim(S,S'_J)$ and the epsilon is changed to $\epsilon_2$. The value of $\epsilon_2$ represents the expected minimum margin ($>>0$) for this criterion. This criterion is evaluated by applying the equation to each sample and plotting a cumulative histogram where the x-axis represents the expected minimum difference margins ranges between [-0.3 to 0.3], and the y-axis represents the number of sentence $S$ that is within the minimum margin $\epsilon_2$.
    
    Further, Figure \ref{fig:jumbling} is a result of swapping $n=3$ words, indicating that the sentence encoders failed to capture the impact of jumbled words on the sentence similarity task. The encoder models generated almost similar semantic representation for the original sentence $S$ and its jumbled version $S'_J$ across all three datasets. This implies that the majority of the samples in figure \ref{fig:jumbling} had a difference between -0.3  to 0, and only a few samples showed positive differences, suggesting that current sentence encoders pay little attention to the word order of the sentence and sheerly rely on contextual words. Therefore, as a result, all five encoders performed in a manner that was opposite to what was expected, resulting in the failure with respect to criterion-3. We have reported the results of all sentence encoders on $n=1$ and $n=2$ in the appendix (omitted here due to lack of space).

\end{enumerate}

\section{Discussions and Conclusion}\label{discussion}

This paper examines the performance of five different sentence encoder models using the SentEval benchmark to evaluate their ability to produce high-quality embeddings for various downstream tasks. The results showed that four out of five models performed reasonably on the benchmark with no single winner for all cases. Although this is a promising result overall, it is still unclear whether the sentence encoders indeed capture basic linguistic properties (which is desired) while performing these downstream tasks or if they are counting on some latent features which are hard to interpret for humans. To further investigate this issue, the paper proposed four criteria to quantify the models' basic semantic understanding abilities in an unsupervised setting and evaluated all five sentence encoding techniques with respect to each criterion.

Our experimental results reveal that the Sentence-BERT model performed the best on the paraphrasing task, while LASER and Infersent were optimal for synonym replacement. However, the experiments conducted on antonym and sentence jumbling tasks revealed limitations in current sentence encoder models' ability to capture desired basic semantic properties. All models failed to differentiate between a sentence and its antonym, as evidenced by the left-skewed cumulative histogram on the respective datasets. The high overlap of words between sentences may be a possible reason for the failure, making it challenging for the encoder to detect subtle differences and leading to similar embeddings. The same results were observed on two additional datasets (QQP and MRPC), confirming the models' inability to capture the semantics of sentences in these tasks.

On the other hand, the evaluation of criterion 4 (Paraphrase Vs. Jumbling) demonstrates that the current sentence encoder models fail to capture the significance of the word order in a sentence; hence, the criterion is not satisfied by any sentence encoder model we tested. Since the current encoders are mainly trained on masked language modeling and next-sentence prediction tasks followed by fine-tuning on many arbitrary downstream tasks, it does not necessarily enforce that the encoders will understand word ordering properly. This is likely to be the reason for the failure of this criterion. Additionally, trends are similar to all three datasets, which confirms the conclusion that current sentence encoders struggle to capture the importance of word order in a sentence.

Based on the above observation, it can be inferred that the current sentence encoder has limitations in capturing the semantic meaning of antonym sentences when there is a high degree of overlapping words. Additionally, the model tends to over-prioritize the contextual words, and as a result, it overlooks the actual ordering of words within the sentence. On the contrary, the same sentence encoders demonstrated high performance on the SentEval benchmark, which consists of several downstream task datasets, despite the limitations discussed above. This raises a daunting dilemma about how a sentence encoder can be considered \textit{``good''} and \textit{``reliable''} when it fails to capture basic linguistic properties like jumbling or antonym interpretation but performs better on downstream tasks. This could be attributed to two possible two reasons, first, either the SentEval benchmark is not hard enough to properly evaluate sentence encoders, or second, the inadequacy of the similarity metrics to assess the sentence relations like cosine-similarity. Alternatively, there is room for improvement in the current sentence encoder models to better capture the semantic meaning of a sentence. Therefore, we believe it is necessary to develop a sentence encoder that can capture the subtle nuances of sentences and generate high-quality embeddings that can reflect all aspects of a sentence.  Additionally, a diverse and robust benchmark is needed to evaluate sentence encoders accurately. Hence, further research and development in this area are needed to improve the capabilities of these models and achieve a more human-like understanding of sentences.

\section{Limitation}\label{limitation}

Our findings are limited to the English language and five popular embedding models (SBert, USE, InferSent, LASER, and Doc2Vec). The experiments are primarily focused on unsupervised semantic understanding tasks where no training data / previous observation about the goal task is available. Thus, evaluation of the constructed perturbed sentences is required.  Therefore, our findings may not hold for all possible downstream NLP tasks. However, in the absence of available training data for a particular domain, our findings can still be very useful to choose a suitable sentence encoder and designing initial experiments.

In future work, we intend to improve upon the limitations discussed above by incorporating the antonyms and word orders to produce more generalized sentence embeddings. Additionally, one can study more recent large language models like chatGPT, and LLaMA to test their limitations on similar criteria.

\bibliographystyle{acl_natbib}
\bibliography{main}
\appendix
\section{Appendix}\label{sec:appendix}

\begin{table*}[!ht]
    \centering
    \begin{tabular}{|c|c|c|c|c|}
    \hline
    \multicolumn{5}{|c|}{\textbf{Average Execution time (in Sec)}} \\ \hline
    \textbf{Model} &\textbf{Paraphrasing} & \textbf{Synonym} & \textbf{Antonym} & \textbf{Jumbling} \\\hline
    \textbf{Universal Sentence Encoder (USE)} &6.01  & 135.7 & 118.6 & 14.3  \\ \hline
    \textbf{Sentence Bert (SBert)} &44.49&131.17&106.40&112.5\\\hline
    \textbf{LASER} & 4.81&14.56 & 15.61&13.74\\\hline
    \textbf{InferSent}& 105.8&386.17 &334.37 &345.14\\\hline
    \textbf{Doc2Vec(D2V)}&108.81& 468.53&411.74&369.23\\\hline
    \end{tabular}
    \caption{Models execution time in secs during hypothesis testing.}
    \label{tab:exec_time}
\end{table*}

\subsection{Hyper-parameter Search}
\begin{enumerate}
    \item \textbf{The exact number of training and evaluation runs : } In this work, our aim was to evaluate each model in zero shot setting. Hence, we used pre-trained models and performed our hypothesis testing. Among all models, only Doc2Vec (D2V) model had undergone for training for 20 epochs, rest all other models were pre-trained and used with their default settings. For \textit{'eps'}, we set values ranging from -0.3 to 0.3.

\end{enumerate}

\subsection{Data-set used}\label{appendix_dataset}
\begin{enumerate}
    \item \textbf{Relevant details such as languages, and number of examples and label distributions: }
    In this work, we experimented with pair of English language sentences. All three datasets used in these works are of different size, so we randomly sampled each dataset to create a balance between all three datasets. The MRPC dataset consist of total 3668 pair of sentences, out of which 1194 pair of sentences having label 0 and other 2474 pair of sentences have label 1. So, to create the balance dataset, we randomly sampled 1194 pairs of sentences from the dataset having label 1. Subsequently, QQP and Paws-WIki dataset has 404290 and 49401 pairs of paraphrased and non-paraphrased sentences. So, we randomly sampled nearly ~1.2K pair of sentences for each labels from each dataset  i.e. ~2.4 pair of sentences collected from each datasets for paraphrasing hypothesis testing. For other hypothesis testing (Synonym and Antonym replacement, Sentence Jumbling), we sampled 3.5K sentences (only single sentence, no pairs of sentence) from all three datasets. Next, we create the perturbed sentences using WordNet toolkit for further hypothesis testing.
    
    \begin{enumerate}[leftmargin=*,itemsep=1ex,partopsep=0ex,parsep=0ex]
\item\textbf{QQP}: This publicly available dataset is the collections of pair of questions on Quora platform \cite{chen2018quora} with labels 1 and 0 annotated by humans. Label 1 is assigned when question 1 and question 2 (one pair) essentially have same meaning (i.e. paraphrased), and otherwise 0 (i.e. non-paraphrased). In this work, we randomly choose ~2.5K samples with label 1 and another ~2.5k samples with label 0  (in total 5k samples) and performed sentence embeddings with different models. 
\item\textbf{PAWS-WIKI}: This publicly available dataset is the collection of pairs of sentences from Wikipedia with high lexical overlapping \cite{zhang2019paws}. In this work, we randomly sampled 5K pairs of sentences out of which 2.5K pairs have label 1 and the rest have label 0. Then, we applied different sentence embedding models and find the similarity between the sentences.  
\item\textbf{MRPC}: The data-set is the collection of pair of sentence collected from newswire articles \cite{dolan-brockett-2005-automatically}. In total, there are ~ 3.5K pair of sentences out of which 1.1K are labeled 0 and the rest are labeled 1 by humans. In this work, we utilize the complete dataset and performed the experiments.
\end{enumerate}
    
    \item \textbf{Details of train/validation/test splits: }
    In this work, we trained our Doc2Vec (D2V) model for 20 epochs. During training, we set the vector size to 100, window size to 5, and minimum  count to 1.
    
    \item \textbf{A zip file containing data or link to a downloadable version of the data : } Yes, we have provided the dataset used in this work with downloadable link. Please follow the readme file in data folder.
    
    \item\textbf{For new data collected, a complete description of the data collection process, such as instructions to annotators and methods for quality control. : } We have create perturbed sentence using the three dataset we used. More details is provided above.  
\end{enumerate}

\subsection{Models Setting}\label{secc:m_para}
All models and hypothesis testing were performed on Google Colab and local system having Intel i5 processor and 8GB of ram. The results/values reported in this work are produced by Colab GPU server.  
\begin{enumerate}
    \item \textbf{USE} \cite{cer2018universal}:  Universal Sentence Encoder (USE) is a transformer-based model that encodes the text to a high fixed 512-dimensional fixed-sized vector. The TF2.0 Saved Model (v4) was load from \cite{tfhub} (thumb). The model has been trained to classify: text classification, sentence similarity, and clustering. 
    
    \item \textbf{SBert} \cite{reimers-gurevych-2019-sentence}: Sentence-BERT is a BERT\cite{devlin-etal-2019-bert} based model which produces semantically meaningful sentence embeddings. In this work, we used \textit{SentenceTransformer} library to load the pre-trained model. and used two pre-trained SBert variants: \textit{"paraphrase-MiniLM-L6-v2"} for hypothesis testing and, \textit{"nli-distilroberta-base-v2"} for other hypothesis testing. The model has been trained on Wikipedia and Book corpus data, and further fine-tuned on the NLI dataset. 
    
    \item \textbf{InferSent} \cite{conneau-etal-2017-supervised}: The model produces sentence embeddings having semantic representations of English sentences. In this work, our model used pre-trained GloVe word embeddings \cite{glove} with 840B tokens, 2.2M vocabulary, 300-dimensional vector, and, InferSent version 1 encoder. We have also set the batch size to 64, word embedding dimension size to 300d, and LSTM encoder size to 2048 with max-pooling layers enabled. Additionally, the model has been trained on the NLI dataset to classify into three categories: entailment, contradiction, and neutral.
    
    \item \textbf{LASER} \cite{laser}: Language-Agnostic-SEntence Representation (LASER) is a model built to perform multilingual sentence embedding tasks and trained in 93 different languages. The model used five BI-LSTM layers in the encoder with max-pooling on the last layer to produce embeddings of a sentence. In this study, we used a pre-trained LASER model with its default settings to produce sentence embedding for a given English sentence. 
    
    \item \textbf{Doc2Vec} \cite{le2014distributed}: Document To Vector (also D2v) learn paragraph and document vector representation as embeddings via the distributed memory and distributed bag of words model. In this work, we set the vector size to 100, the window size to 5, minimum count to 1 with 20 epochs for training. 
\end{enumerate}

\subsection{Results}
\subsubsection{Hyothesis-4: Jumble Sentence}\label{apend_jumbling}
\begin{figure*}[!pt]
    \centering
    \begin{subfigure}{0.8\textwidth}
        \includegraphics[width=\linewidth]{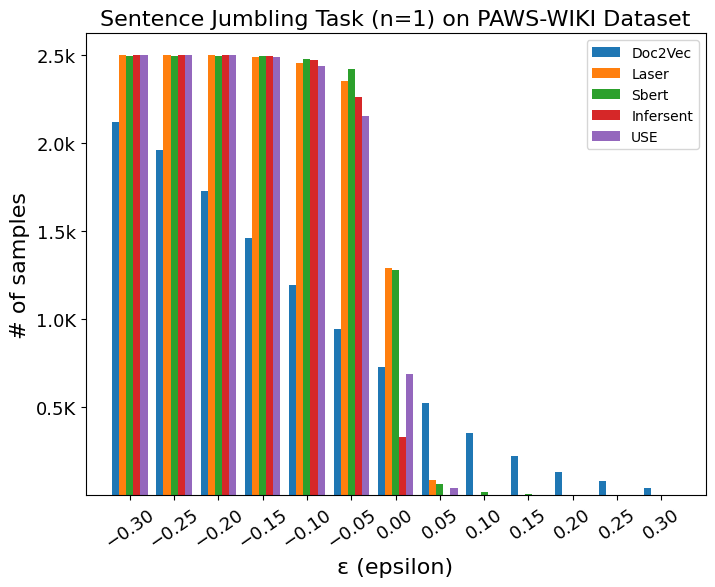}
        \caption{}
        \label{fig:pn1}
    \end{subfigure}
    
    \begin{subfigure}{0.8\textwidth}
        \includegraphics[width=\linewidth]{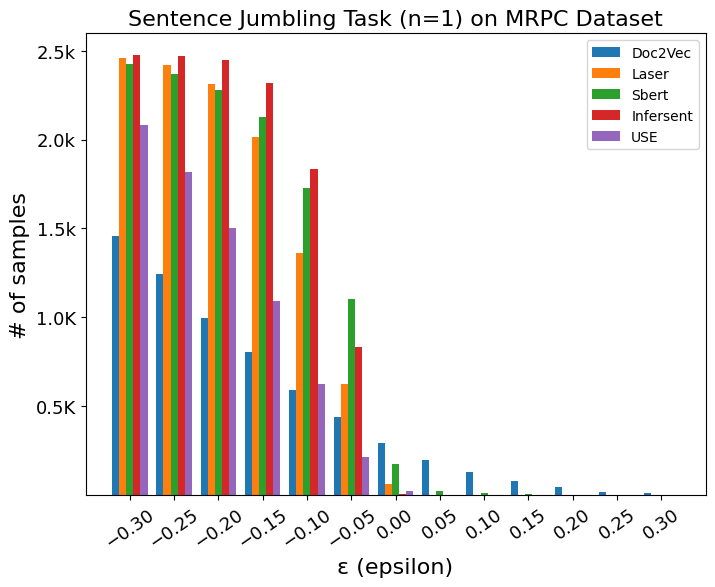}
        \caption{}
        \label{fig:mn1}
    \end{subfigure}
    
    \caption{The figures indicate the Jumble Sentence Hypothesis's cumulative histogram for n=1. The scores are determined by the equation $Sim(S,S'_P) - Sim(S,S_J') > \epsilon_2$, where $\epsilon_2$ represents the expected minimum margin. The performance of five different sentence encoders is tested on three datasets.}
    \label{fig:jumbling_n1}
\end{figure*}

\begin{figure*}[!hb]
        \centering
        \includegraphics[width =0.8\textwidth]{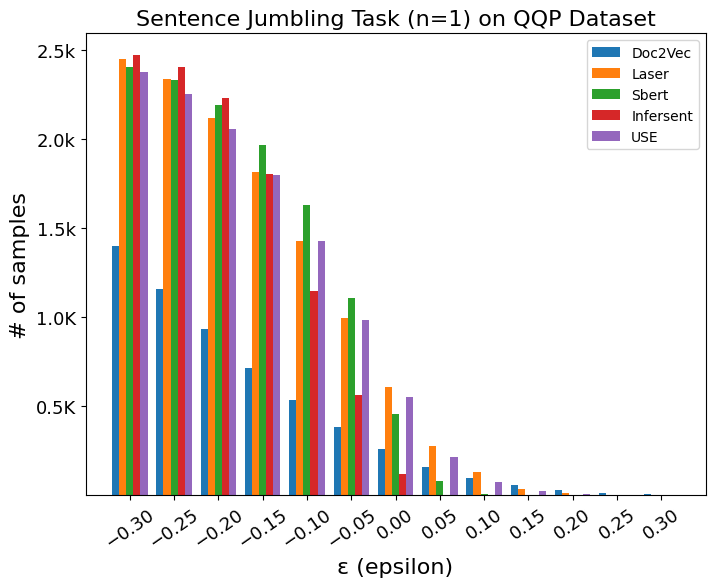}
        \caption{The figures indicate the Jumble Sentence Hypothesis's cumulative histogram for n=1 for the PAWS-WIKI dataset. The scores are determined by the equation $Sim(S,S'_P) - Sim(S,S_J') > \epsilon_2$, where $\epsilon_2$ represents the expected minimum margin. The performance of five different sentence encoders is tested on three datasets.}
        \label{fig:jump_paws_n1_2}
    \end{figure*}

\begin{figure*}[!hb]
        \centering
        \includegraphics[width =0.8\textwidth]{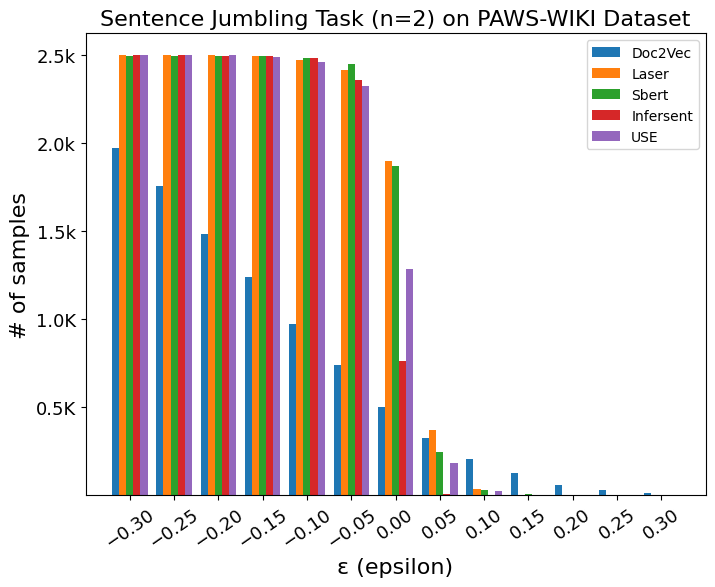}
        \caption{The figures indicate the Jumble Sentence Hypothesis's cumulative histogram for n=1 for the PAWS-WIKI dataset. The scores are determined by the equation $Sim(S,S'_P) - Sim(S,S_J') > \epsilon_2$, where $\epsilon_2$ represents the expected minimum margin. The performance of five different sentence encoders is tested on three datasets.}
        \label{fig:jump_paws_n2_2}
    \end{figure*}

\begin{figure*}[!pt]
    \centering
    \begin{subfigure}[b]{0.8\textwidth}
        \includegraphics[width=\textwidth]{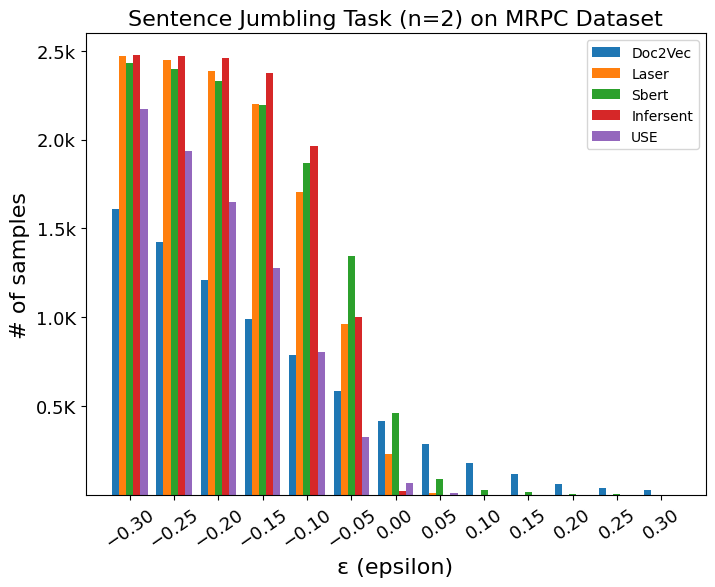}
        \caption{}
        \label{fig:jm2}
    \end{subfigure}
    \begin{subfigure}[b]{0.8\textwidth}
        \includegraphics[width=\textwidth]{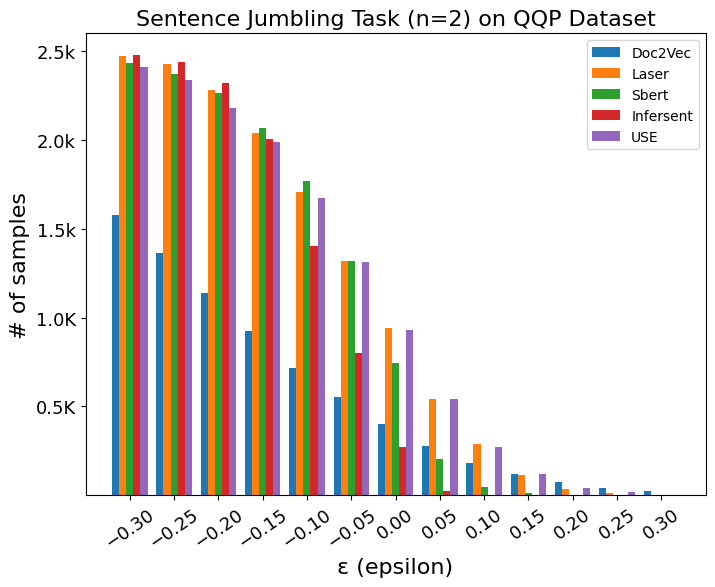}
        \caption{}
        \label{fig:jq2}
    \end{subfigure}
    \caption{The figures indicate the Jumble Sentence Hypothesis's cumulative histogram for n=2. The scores are determined by the equation $Sim(S,S'_P) - Sim(S,S_J') > \epsilon_2$, where $\epsilon_2$ represents the expected minimum margin. The performance of five different sentence encoders is tested on three datasets.}
    \label{fig:jumbling_n2}
\end{figure*}

The results of the cosine similarity difference for the Jumble Sentence task are shown in Figure [\ref{fig:jumbling_n1},\ref{fig:jump_paws_n1_2},\ref{fig:jump_paws_n2_2} and, \ref{fig:jumbling_n2} ]. The all figures showcase the models ability to capture semantic meaning when the words are swapped by order of \textit{'n'} i.e. n={1,2,3} across all three datasets. The difference score is calculated as $Sim(S,S'_P) - Sim(S,S'_J) > \epsilon_2$. All five sentence encoders were evaluated on three datasets, and the results suggest that the models struggle to capture the word order of sentences and, therefore, generate poor sentence embeddings. The figure displays the cumulative number of samples with a difference in cosine similarity score greater than $\epsilon_2$.

s

\end{document}